\documentclass[runningheads]{llncs}
\usepackage{times}
\usepackage{latexsym}
\usepackage{color}
\usepackage{oldlfont,rawfonts}
\usepackage{graphicx}
\usepackage{verbatim}

\usepackage{amssymb}
\usepackage{amsmath}

\usepackage{colortbl}
\usepackage{microtype}

\newcommand {\rf} {\mathit{rank}}

\newcommand {\lingconc} {\mathcal{S}}

\newcommand {\ent} {\mathrel{{\scriptstyle\mid\!\sim}}}
\newcommand {\imp} {\rightarrow}

\newcommand {\sx} {\langle}
\newcommand {\dx} {\rangle}

\newcommand {\emme} {{\mathcal{M}}}
\newcommand {\enne} {\mathcal{N}}

\newcommand {\tc} {\mid}
\newcommand {\vuoto} {\emptyset}

\newcommand{\tip}{{\bf T}}

\newcommand{\alc}{\mathcal{ALC}}

\newcommand{\alct}{\mathcal{ALC}+\tip}
\newcommand{\alctmin}{{\mathcal{ALC}}+\tip_{min}}

\newcommand{\elpb}{{\mathcal{EL}}^{+}_{\bot}}

\newcommand{\be}{\begin{enumerate}}
\newcommand{\ee}{\end{enumerate}}

\newcommand{\hide}[1]{}

\def \cases{\left \{\begin{array}{l}}
\def \endcases{\end{array}\right .}

\newcommand {\Pe} {{\bf P}}

\newcommand {\bes} {\begin{description}}
\newcommand{\ens} {\end{description}}

\newcommand {\beq} {\begin{quote}}
\newcommand {\enq} {\end{quote}}
\newcommand {\bit} {\begin{itemize}}
\newcommand {\enit} {\end{itemize}}

\newenvironment{pozz}{\color{black}}{\color{black}}

\begin{document}
\bibliographystyle{plain}

 \title
 { 
A framework for a modular multi-concept  \\
lexicographic closure semantics 
  }
  

\author{Laura Giordano \and  Daniele Theseider Dupr{\'{e}} }

\institute{DISIT - Universit\`a del Piemonte Orientale, Italy\\
 \email{laura.giordano@uniupo.it, dtd@uniupo.it}
}


\maketitle
 
 \begin{abstract} 
 We define a  modular multi-concept extension of the lexicographic closure semantics for defeasible description logics with typicality. The idea is that of distributing the defeasible properties of concepts into different modules, according to their subject, and of defining a notion of preference for each module based on the  lexicographic closure semantics.  The preferential semantics of the knowledge base can then be defined as a combination of the preferences of the single modules. The range of possibilities, from fine grained to coarse grained modules, provides a spectrum of alternative semantics.

\end{abstract}


\vspace{-0.1cm}
\section{Introduction}
\vspace{-0.1cm}

Kraus, Lehmann and Magidor's preferential logics for non-monotonic reasoning  \cite{KrausLehmannMagidor:90,whatdoes}, 
have been extended to description logics, to deal with inheritance with exceptions in ontologies,
allowing for non-strict forms of inclusions,
called {\em typicality or defeasible inclusions}, with different preferential and ranked semantics \cite{lpar2007,sudafricaniKR} 
as well as different closure constructions such as the rational closure \cite{casinistraccia2010,CasiniDL2013,dl2013,AIJ15},  
the lexicographic closure \cite{Casinistraccia2012}, the relevant closure \cite{Casini2014}, and MP-closure \cite{Ecsqaru19}.



In this paper we define a modular multi-concept extension of the lexicographic closure for reasoning about exceptions in ontologies. The idea is very simple: different modules can be defined starting from a defeasible knowledge base, containing a set ${\cal D}$ of typicality inclusions (or defeasible inclusions) describing the prototypical properties of  classes in the knowledge base. 
We will represent such defeasible inclusions as $\tip(C) \sqsubseteq D$ \cite{lpar2007}, meaning that ``typical $C$'s are $D$'s" or ``normally $C$'s are $D$'s",
corresponding to conditionals $C \ent D$ in KLM framework. 

A set of modules $m_1, \ldots, m_n$ is introduced,  each one concerning a subject, 
and defeasible inclusions belong to a module if they are related 
with  its subject.  By subject, here, we mean any concept of the knowledge base.
 Module $m_i$ with subject $C_i$ does not need to contain just typicality inclusions of the form $\tip(C_i) \sqsubseteq D$, but all defeasible inclusions in ${\cal D}$ which are concerned with subject $C_i$ are admitted in $m_i$. 
We call a collection of such modules a {\em modular multi-concept knowledge base}.
 
This modularization of the defeasible part of the knowledge base does not define a partition of the set ${\cal D}$ of defeasible inclusions, as an inclusion may belong to more than one module. For instance, the typical properties of employed students are relevant both for the module with subject  $\mathit{Student}$ and  for the  module with subject  $\mathit{Employee}$.
The granularity of modularization has to be chosen by the knowledge engineer who can fix how large or narrow is the scope of a module, and how many modules are to be included in the knowledge base (for instance, whether the properties of employees and students are to be defined in the same module with subject $\mathit{Person}$ or in two different modules). 
At one extreme, all the defeasible inclusions in ${\cal D}$ can be put together in a module associated with subject $\top$ (Thing). At the other extreme,
which has been studied in \cite{arXiv_ICLP2020}, 
a module $m_i$ is a defeasible TBox  containing {\em only} the defeasible inclusions of the form  $\tip(C_j) \sqsubseteq D$ for some concept $C_i$. 
In this paper we remove this restriction considering general modules, containing arbitrary sets of defeasible inclusions, intuitively pertaining 
 some subject.

In \cite{arXiv_ICLP2020}, following Gerard Brewka's framework of Basic Preference Descriptions  for ranked knowledge bases \cite{Brewka04},
 we have assumed that a specification of the relative importance of typicality inclusions for a concept $C_i$ is given by assigning ranks to typicality inclusions. However, 
for a large module, a specification by hand of the ranking of the defeasible inclusions in the module would be awkward. 
In particular, a module may include all  properties of a class  as well as  properties of its exceptional subclasses (for instance, the typical properties of penguins, ostriches, etc.  might  all be included in a module with subject $\mathit{Bird}$).
A natural choice is then to consider, for each module, a lexicographic semantics which builds on the rational closure ranking  to define a preference ordering on domain elements. This preference relation corresponds, in the propositional case, to the lexicographic order on worlds in Lehmann's model theoretic semantics of the lexicographic closure \cite{Lehmann95}. 
This semantics already accounts for the specificity relations among concepts inside the module, as the lexicographic closure deals with specificity, based on  ranking of concepts computed by the rational closure of the knowledge base.


Based on the ranked semantics of the single modules,
a compositional (preferential) semantics of 
the knowledge base is defined 
by combining the multiple preference relations into a single global preference relation $<$.
%
This gives rise to a modular multi-concept extension of Lehmann's preference semantics for the lexicographic closure.
When there is a single module, 
containing all the typicality inclusions in the knowledge base, the semantics collapses to a natural extension to DLs of Lehmann's semantics, 
which corresponds to Lehmann's semantics 
for the fragment of $\alc$ without universal and existential restrictions. 

We introduce a  notion of  entailment for modular multi-concept knowledge bases, 
based on the proposed semantics, which satisfies the KLM properties of a preferential consequence relation. 
This notion of entailment has  good properties inherited from lexicographic closure:
it deals properly with irrelevance and specificity, and it is not subject to the ``blockage of property inheritance" problem,  i.e., the problem that property inheritance from classes to subclasses is not guaranteed,
which affects the rational closure \cite{PearlTARK90}.
In addition, separating defeasible inclusions in different modules 
provides a simple solution to another problem of the rational closure and its refinements 
(including the lexicographic closure), 
that was recognized by Geffner and Pearl \cite{GeffnerAIJ1992}, namely, that ``conflicts among defaults that should remain unresolved, are resolved anomalously", giving rise to too strong conclusions.
The preferential (not necessarily ranked) nature of the global preference relation $<$ provides a simple way out to this problem, when defeasible inclusions are suitably separated in different modules. 

\section{Preliminaries: The description logics $\alc$ and its extension with typicality inclusions}\label{sec:ALC}

Let ${N_C}$ be a set of concept names, ${N_R}$ a set of role names
  and ${N_I}$ a set of individual names.  
The set  of $\alc$ \emph{concepts} (or, simply, concepts) can be
defined inductively as follows:

 \begin{itemize}
\item
$A \in N_C$, $\top$ and $\bot$ are {concepts};
    
\item
if $C$ and $ D$ are concepts and $R \in N_R$, then $C
\sqcap D, C \sqcup D, \neg C, \forall R.C, \exists R.C$ are
{concepts}.
\end{itemize}


\noindent  
A knowledge base (KB) $K$ is a pair $({\cal T}, {\cal A})$, where ${\cal T}$ is a TBox and
${\cal A}$ is an ABox.
The TBox ${\cal T}$ is  a set of concept inclusions (or subsumptions) $C \sqsubseteq D$, where $C,D$ are concepts.
The  ABox ${\cal A}$ is  a set of assertions of the form $C(a)$ 
and $R(a,b)$ where $C$ is a  concept, $R \in
N_R$, and $a, b \in N_I$.

An  $\alc$ {\em interpretation}  \cite{handbook} is a pair $I=\langle \Delta, \cdot^I \rangle$ where:
$\Delta$ is a domain---a set whose elements are denoted by $x, y, z, \dots$---and 
$\cdot^I$ is an extension function that maps each
concept name $C\in N_C$ to a set $C^I \subseteq  \Delta$, each role name $R \in N_R$
to  a binary relation $R^I \subseteq  \Delta \times  \Delta$,
and each individual name $a\in N_I$ to an element $a^I \in  \Delta$.
It is extended to complex concepts  as follows:
\begin{align*}
&\top^I  =\Delta     \;\;\;\;\;\;     \bot^I =\vuoto \\
	  & (\neg C)^I=\Delta \backslash C^I \\
	 &(C \sqcap D)^I  =C^I \cap D^I  \\
	 & (C \sqcup D)^I =C^I \cup D^I\\
	&(\forall R.C)^I =\{x \in \Delta \tc \forall y. (x,y) \in R^I \imp y \in C^I\} \\ 
	&(\exists R.C)^I =\{x \in \Delta \tc \exists y.(x,y) \in R^I \ \& \ y \in C^I\}.
\end{align*}
%
%
%
%
The notion of satisfiability of a KB  in an interpretation and the notion of entailment are defined as follows:

\begin{definition}[Satisfiability and entailment] \label{satisfiability}
Given an $\alc$ interpretation $I=\langle \Delta, \cdot^I \rangle$: 

\begin{quote}
	- $I$  satisfies an inclusion $C \sqsubseteq D$ if   $C^I \subseteq D^I$;
	
	-   $I$ satisfies an assertion $C(a)$ if $a^I \in C^I$; 
	
	-   $I$ satisfies an assertion $R(a,b)$ if $(a^I,b^I) \in R^I$.
\end{quote}

\noindent
 Given  a KB $K=({\cal T}, {\cal A})$, 
 an interpretation $I$  satisfies ${\cal T}$ (resp., ${\cal A}$) if $I$ satisfies all  inclusions in ${\cal T}$ (resp., all assertions in ${\cal A}$).
 $I$ is an $\alc$ \emph{model} of $K=({\cal T}, {\cal A})$ if $I$ satisfies ${\cal T}$ and ${\cal A}$.

 Letting a {\em query} $F$ to be either an inclusion $C \sqsubseteq D$ (where $C$ and $D$ are concepts) 
or an assertion ($C(a)$ or $R(a,b)$), 
 {\em $F$ is entailed by $K$}, written $K \models_{\alc} F$, if for all $\alc$ models $I=$$\sx \Delta,  \cdot^I\dx$ of $K$,
$I$ satisfies $F$.
\end{definition}
Given a knowledge base $K$,
the {\em subsumption} problem is the problem of deciding whether an inclusion $C \sqsubseteq D$ is entailed by  $K$.
The {\em instance checking} problem is the problem of deciding whether an assertion $C(a)$ is entailed by $K$.
The {\em concept satisfiability} problem is the problem of deciding, for a concept $C$, whether $C$ is consistent with $K$ (i.e., whether there exists a model $I$ of $K$, such that $C^I \neq \emptyset$).

In the following we will refer to an extension of $\alc$ with typicality inclusions, that we will call $\alct$ as in \cite{lpar2007}, and to
the {\em rational closure} of $\alct$ knowledge bases $({\cal T}, {\cal A})$ 
\cite{dl2013,AIJ15}. In addition to standard $\alc$ inclusions $C \sqsubseteq D$ (called  {\em strict} inclusions in the following), in $\alct$ the TBox ${\cal T}$ also contains typicality inclusions of the form $\tip(C) \sqsubseteq D$, where $C$ and $D$ are $\alc$ concepts. Among all rational closure constructions for $\alc$ mentioned in the introduction, we will refer to the one in \cite{dl2013}, and to its minimal canonical model semantics. 
Let us recall the notions of preferential, ranked and canonical model of a defeasible knowledge base $({\cal T}, {\cal A})$, that will be useful in the following.

\begin{definition}[Interpretations for $\alct$]\label{semalctr} 
A {\em preferential} interpretation $\enne$ is any
structure $\langle \Delta, <, \cdot^I \rangle$ where: $\Delta$ is a
domain;   
 $<$ is an irreflexive, transitive and well-founded 
 relation over $\Delta$;
 $\cdot^I$ is a function that maps all concept names, role names and individual names as defined above for $\alc$ interpretations,
 and provides an interpretation to all $\alc$ concepts as above, and to typicality concepts as follows:
$(\tip(C))^I = min_<(C^I)$, 
where
$min_<(S)= \{u: u \in S$ and $\nexists z \in S$ s.t. $z < u \}$.

\noindent
When  relation $<$ is  required to be also modular 
(i.e., for all $x,y,z \in \Delta$, if $x < y$ then $x < z$ or $z < y$), $\enne$ is called a {\em ranked} interpretation.
\end{definition}
Preferential interpretations for description logics were first studied in \cite{lpar2007}, while ranked interpretations (i.e., modular preferential interpretations)  were first introduced for $\alc$ in \cite{sudafricaniKR}. 

A preferential (ranked) model of an $\alct$ knowledge base $K$ is a preferential (ranked) $\alct$ interpretation ${\enne}=\langle \Delta, <, \cdot^I \rangle$ that satisfies all inclusions in $K$, where:  
a strict inclusion or an assertion is satisfied in $\enne$ if  it is satisfied in the $\alc$ model $\langle \Delta, \cdot^I \rangle$,
and a typicality  inclusion $\tip(C) \sqsubseteq D$ is satisfied in $\enne$ if   $(\tip(C))^I \subseteq D^I$.

\noindent
Preferential entailment in $\alct$ is defined in the usual way: for a knowledge base $K$ and a query $F$ (a strict or defeasible inclusion or an assertion),  $F$ is {\em preferentially entailed} by $K$ ($K \models_{\alct} F$) if $F$ is satisfied in all preferential models of $K$. 

A canonical model for $K$ is a preferential (ranked) model containing, roughly speaking, as many domain elements as consistent with the knowledge base specification $K$.
Given an  $\alct$ knowledge base $K= ({\cal T}, {\cal A})$ and a query $F$,
let us define $\lingconc_K$ as the set of all $\alc$ concepts (and subconcepts) occurring  in $K$ or in $F$,
together with their complements. 
We consider all the {\em sets of  concepts $\{C_1, C_2, \dots,$ $ C_n\} \subseteq \lingconc_K$  consistent with $K$}, i.e., s.t. $K \not\models_{\alct} C_1 \sqcap C_2 \sqcap \dots \sqcap C_n \sqsubseteq \bot$.

\begin{definition}[Canonical model] \label{def-canonical-model-DL}. 
A  preferential model $\emme=$$\sx \Delta, <, I \dx$ of $K$ is 
{\em canonical with respect to $\lingconc_K$} if it contains at least a domain element $x \in \Delta$ s.t.
$x \in (C_1 \sqcap C_2 \sqcap \dots \sqcap C_n)^I$, for each set
$\{C_1, C_2, \dots, C_n\} \subseteq \lingconc_K$ consistent with $K$. 
\end{definition}
For finite, consistent  $\alct$ knowledge bases, existence of finite (ranked) canonical models has been proved in \cite{AIJ15} (Theorem 1).
In the following, as we will only consider finite $\alct$ knowledge bases,  we can restrict our consideration to {\em finite} preferential models.


\section{Modular multi-concept knowledge bases}\label{sez:MC_KB}

In this section we introduce a notion of a multi-concept knowledge base, starting from
 a set of strict inclusions ${\cal T}$, a set of assertions ${\cal A}$, and a set of typicality inclusions ${\cal D}$, each one of the form $\tip(C) \sqsubseteq D$, where $C$ and $D$ are $\alc$ concepts.
\begin{definition}
A {\em modular multi-concept knowledge base}  $K$ is a tuple  $\mathit{\langle {\cal T}, {\cal D}, m_1, \ldots, }$ $\mathit{m_k, {\cal A}, s \rangle}$, where
$ {\cal T}$  is an $\alc$ TBox,  ${\cal D}$ is a set of typicality inclusions, 
such that $m_1 \cup \ldots \cup m_k = {\cal D}$,
${\cal A}$ is an ABox, and  $s$ is a function associating each module $m_i$ with a concept, $s(m_i)= C_i $, the { \em subject} of  $m_i$.
\end{definition}
The idea is that each $m_i$ is a {\em module} defining the typical properties of the instances of some concept $C_i$. 
The defeasible inclusions belonging to a module $m_i$ with subject $C_i$ are the inclusions that intuitively pertain to $C_i$.
We expect that all the typicality inclusions $\tip(C) \sqsubseteq D$, such that $C$ is a subclass of $C_i$, belong to $m_i$, but not only.
For instance, for a module $m_i$ with subject  $C_i=\mathit{Bird}$, the typicality inclusion 
$\mathit{\tip(Bird \sqcap Live\_at\_SouthPole) \sqsubseteq Penguin}$, meaning  that the birds living at the south pole are normally penguins, is clearly to be included in $m_i$. As penguins are birds, also  inclusion $\mathit{\tip(Penguin)  \sqsubseteq Black}$ 
is to be included in $m_i$,
and, if  $\mathit{\tip(Bird)  \sqsubseteq Flying}$- $\mathit{Animal}$ and $\mathit{\tip(FlyingAnimal)  \sqsubseteq BigWings}$ are defeasible inclusions in the knowledge base, they both  may be relevant properties of birds to be included in $m_i$.
For this reason we will not put 
restrictions on the typicality inclusions that can belong to a module. 
We will see later that the semantic construction for a module $m_i$ will be able to ignore the typicality inclusions which are not relevant for subject $C_i$
and that there are cases when not even the inclusions $\tip(C) \sqsubseteq D$ with $C$ subsumed by $C_i$ are admitted in $m_i$.

The modularization $m_1, \ldots, m_k$ of the defeasible part ${\cal D}$ of the knowledge base  does not define a partition of ${\cal D}$, as the same inclusion may belong to more than one module $m_i$. For instance, the typical properties of employed students are relevant   for both concept  $\mathit{Student}$ and  concept  $\mathit{Employee}$ and should belong to their related modules (if any).
Also, a granularity of modularization has to be chosen and, as we will see, this choice may have an impact on the global semantics of the knowledge base.
At one extreme, all the defeasible inclusions in ${\cal D}$ are put together in the same module, e.g., the module associated with concept $\top$. At the other extreme,
which has been studied in \cite{arXiv_ICLP2020}, 
a module $m_i$ contains {\em only} the defeasible inclusions of the form  $\tip(C_i) \sqsubseteq D$, where $C_i$ is the subject of $m_i$
(and in this case, the inclusions $\tip(C) \sqsubseteq D$ with $C$ subsumed by $C_i$ are not admitted in $m_i$). 
%
In this regard, the framework proposed in this paper could be seen as an extension of the proposal in  \cite{arXiv_ICLP2020} 
to allow coarser grained modules, 
while here we do not allow for user-defined preferences among defaults.

Let us consider an example of multi-concept knowledge base.

\begin{example} \label{exa:student}
Let $K$ be the  knowledge base
$\mathit{\langle {\cal T}, {\cal D}, m_1, m_2, m_3, {\cal A}, s \rangle}$,
where ${\cal A} =\emptyset$, 
${\cal T}$ contains the strict inclusions:

\begin{quote}
$\mathit{Employee  \sqsubseteq  Adult}$ \ \ \ \ \ \ \ \ \ \ \ \ \ \ \ \ \ \ \  \\
$\mathit{Adult  \sqsubseteq  \exists has\_SSN. \top}$  \\
$\mathit{PhdStudent  \sqsubseteq  Student}$\  \ \ \ \ \ \ \ \ \ \ \ \  \\
$\mathit{PhDStudent \sqsubseteq  Adult}$ \\
$\mathit{   Has\_no\_Scolarship \equiv \neg  \exists hasScolarship.\top}$ \ \ \ \ \ \ \ \ \ \ \ \  $\mathit{PrimarySchoolStudent \sqsubseteq  Children}$ \\
$\mathit{PrimarySchoolStudent \sqsubseteq  HasNoClasses}$ 
$\mathit{Driver \sqsubseteq  Adult}$ \ \ \ \ \ \ \ \ \ \ \ \  \ \ \ \ \\
$\mathit{Driver \sqsubseteq  \exists has\_DrivingLicence. \top}$
\end{quote}
and the defeasible inclusions in ${\cal D}$ are distributed in the modules $m_1, m_2, m_3$ as follows.

Module $m_1$ has subject  $Employee$, and contains the defeasible inclusions:

\noindent
$(d_1)$ $\mathit{\tip(Employee) \sqsubseteq \neg Young}$ \\ 
$(d_2)$ $\mathit{\tip(Employee) \sqsubseteq \exists has\_boss.Employee}$ \\
$(d_3)$ $\mathit{\tip(ForeignerEmployee) \sqsubseteq \exists has\_Visa.\top}$\\
$(d_4)$  $\mathit{\tip(Employee \sqcap Student) \sqsubseteq Busy}$\\
$(d_5)$  $\mathit{\tip(Employee \sqcap Student) \sqsubseteq  \neg Young}$

Module $m_2$ has subject  $Student$, and contains the defeasible inclusions:

\noindent
$(d_6)$ $\mathit{\tip(Student) \sqsubseteq  \exists has\_classes.\top}$\\ 
$(d_7)$ $\mathit{\tip(Student) \sqsubseteq Young}$\\
$(d_8)$ $\mathit{\tip(Student) \sqsubseteq  Has\_no\_Scolarship}$\\
$(d_9)$ $\mathit{\tip(HighSchoolStudent) \sqsubseteq  Teenager}$\\
$(d_{10})$ $\mathit{\tip(PhDStudent) \sqsubseteq  \exists hasScolarship.Amount}$ \\ 
$(d_{11})$ $\mathit{\tip(PhDStudent) \sqsubseteq Bright}$\\
$(d_4)$  $\mathit{\tip(Employee \sqcap Student) \sqsubseteq Busy}$\\
$(d_5)$  $\mathit{\tip(Employee \sqcap Student) \sqsubseteq  \neg Young}$

%
%
%

Module $m_3$ has subject  $Vehicle$, and contains the defeasible inclusions:

\noindent
$(d_{12})$ $\mathit{\tip(Vehicle) \sqsubseteq  \exists has\_owner.Driver}$\\ 
$(d_{13})$ $\mathit{\tip(Car) \sqsubseteq  \neg SportsCar}$\\ 
$(d_{14})$ $\mathit{\tip(SportsCar) \sqsubseteq RunFast}$\\
$(d_{15})$ $\mathit{\tip(Truck) \sqsubseteq Heavy}$\\
$(d_{16})$ $\mathit{\tip(Bicycle) \sqsubseteq \neg RunFast}$



%
%
%

\end{example}
Observe that, in previous example, 
$(d_4)$ and $(d_5)$ belong to both modules $m_1$ and $m_2$.
An additional module  might be added containing the prototypical properties of Adults.

\section{A lexicographic semantics of modular multi-concept knowledge bases} \label{sec:semantics}

In this section, we define a semantics of modular multi-concept knowledge bases, based on Lehmann's lexicographic closure semantics \cite{Lehmann95}.
The idea is that, for each module $m_i$, a semantics can be defined using lexicographic closure semantics, with some minor modification.  

Given a modular multi-concept knowledge base  $K= \mathit{\langle {\cal T}, {\cal D}, m_1, \ldots, m_k, {\cal A}, s \rangle}$,
we let $\rf(C)$ be the rank of concept $C$ in the rational closure ranking of the knowledge base $({\cal T}\cup {\cal D}, {\cal A})$, according to the {\em rational closure} construction in \cite{dl2013}. 
In the rational closure ranking, concepts with higher ranks are more specific than concepts with lower ranks.
While we will not recall the rational closure construction, let us consider again Example \ref{exa:student}.
In Example \ref{exa:student}, the rational closure ranking assigns to
concepts $\mathit{Adult}$, $\mathit{Employee}$, $\mathit{ForeignEmployee}$, $\mathit{Driver}$, $\mathit{Student}$, $\mathit{HighSchoolStudent}$, $\mathit{Primary}$- \linebreak  $\mathit{SchoolStudent}$ the rank $0$, while to concepts
$\mathit{PhDStudent}$ and $\mathit{Employee \sqcap Student}$ the rank $1$. In fact, $\mathit{PhDStudent}$ are exceptional students, as they have a scholarship, while employed students are exceptional students, as they are not young. Their rank is higher than the rank of concept $\mathit{Student}$ as they are exceptional subclasses of class $\mathit{Student}$.

Based on the concept ranking, the rational closure assigns a rank to typicality inclusions: the rank of $\tip(C) \sqsubseteq D$ is equal to the rank of  concept $C$.
For each module $m_i$ of a knowledge base $K=\mathit{\langle {\cal T}, {\cal D}, m_1, \ldots, m_k, {\cal A}, s \rangle}$,
we aim to define a canonical model, using the lexicographic order based on the rank of typicality inclusions in $m_i$. 
In the following we will assume that the knowledge base $\langle {\cal T}\cup {\cal D}, {\cal A}\rangle$ is consistent in the logic $\alct$, that is, it has a preferential model. This also guarantees the existence of  (finite) canonical models \cite{AIJ15}.
In the following, as the knowledge base $K$ is finite,  we will restrict our consideration to {\em finite} preferential and ranked models.

Let us define the {\em projection of the knowledge base $K$ on module $m_i$} as the knowledge base $K_i= \langle {\cal T}\cup m_i, {\cal A}\rangle$.
$K_i$ is an $\alct$ knowledge base.
Hence a preferential model ${\enne}_i= \langle \Delta, <_i, \cdot^I \rangle$  of $K_i$
is defined as  in Section \ref{sec:ALC}
(but now we use $<_i$, instead of $<$, for the preference relation in ${\enne}_i$, for $i=1,\ldots, k$).


In his seminal work on the lexicographic closure, 
Lehmann \cite{Lehmann95}
defines a model theoretic semantics of the lexicographic closure construction by introducing an order relation among propositional models, 
considering which defaults are violated in each model, and introducing a
seriousness ordering $\prec$ among sets of violated defaults.
For two propositional models $w$ and $w'$, $w \prec w' $ ($w$ is preferred to $w'$) is defined in \cite{Lehmann95} as follows:
%
\begin{align}\label{cond:preference_prop_int}
w \prec w' \mbox{\ \ {\bf \em iff} } V(w)  \prec V(w')
\end{align}
$w$ is preferred to $w'$ when the defaults $V(w)$ violated by $w$ are less serious than  the defaults $V(w')$ violated by $w'$.
As we will recall below, the seriousness ordering depends on the number of defaults violated by $w$ and by $w'$ for each rank.
 
In a similar way, in the following, we introduce a ranked relation $<_i$ on the domain $\Delta$ of a  model of $K_i$.
Let us first define, 
for a preferential model ${\enne}_i= \langle \Delta, <_i, \cdot^I \rangle$ of $K_i$, 
what it means that an element $x \in \Delta$ {\em violates} a typicality inclusion $\tip(C) \sqsubseteq D$ in $m_i$.
\begin{definition} \label{def:violation}
Given a module $m_i$ of $K$, with  $s(m_i)=C_i$, and  
a preferential model ${\enne}_i= \langle \Delta, <_i, \cdot^I \rangle$ of $K_i$, 
an element $x \in \Delta$ {\em violates} a typicality inclusion $\tip(C) \sqsubseteq D$ in $m_i$ if $x \in C_i^I$, $x \in C^I$ and $x \not \in D^I$.
\end{definition}
Notice that, the set of typicality inclusions violated by a domain element $x$ in a model only depends on the interpretation  $\cdot^I$ of $\alc$ concepts, and on the defeasible inclusions in $m_i$.
Furthermore, differently from the usual notion of violation in Lehmann's semantics, for a module $m_i$ with subject $C_i$, we do not consider the violations of domain elements  $x \not  \in C_i^I$ (i.e., the domain elements $x$ which are not $C_i$-instances are assumed not to violate any default in $m_i$).
Let $V_i(x)$ be the set of the defeasible inclusions of $m_i$ violated by domain element $x$, and let $V_i^h(x)$ be the set of all defeasible inclusions in $m_i$ with rank $h$ which are violated by domain element $x$. 

In order to compare alternative sets of defaults, in  \cite{Lehmann95}  the seriousness ordering $\prec$ among sets of defaults is defined by associating with each set of defaults $D \subseteq K$ 
a tuple of numbers $\langle n_0, n_1, \ldots, n_r \rangle$, where 
$r$ is the {\em order} of $K$, i.e. the least finite $i$ such that 
there is no default with the finite rank $r$ or  rank higher than $r$ (but there is at least one default with rank $r-1$). The tuple is constructed considering the ranks of defaults in the rational closure.
 $n_0$ is the number of defaults in $D$ with rank $\infty$ 
and, for $1 \leq i \leq  k$,  $n_i$ is the number of defaults  in $D$ with rank $r-i$ (in particular, $n_r$ is the number of defaults in $D$ with rank $0$).
Lehmann defines the strict modular order $\prec$ among sets of defaults  from the natural lexicographic order over the tuples $\langle n_0, n_1, \ldots, n_k \rangle$. 
This order gives preference to those sets of defaults containing a larger number of more specific defaults. As we have seen from equation  (\ref{cond:preference_prop_int}), $\prec$ is used by Lehmann to compare sets of violated defaults and to prefer the propositional models whose violations are less serious. 

We use the same criterion for comparing domain elements, introducing a seriousness ordering $\prec_i$ for each module $m_i$.
Considering that the defaults with infinite rank must be satisfied by all domain elements,  we will not need to consider their violation in our definition (that is, we will not consider $n_0$ in the following).

The set  $V_i(x)$ of defaults from module $m_i$ which are violated by $x$, can be associated with a tuple of numbers $t_{i,x}= \langle |V_i^{r-1}(x)|, \ldots, |V_i^0(x)| \rangle$.
Following Lehmann, we let $V_i(x)  \prec_i V_i(y)$  iff $t_{i,x}$ comes before $t_{i,y}$ 
in the natural lexicographic order on tuples (restricted to the violations of defaults in $m_i$), that is:
\begin{align*}
V_i(x)  \prec_i V_i(y)   \mbox{  \ \ iff \ \ }  & \mbox{ $\exists l$ such that  $|V_i^l(x)| < |V_i^l(y)|$} \\
& \mbox{  and, $\forall h>l$,  $|V_i^h(x)| = |V_i^h(y)|$}
\end{align*}

\begin{definition}\label{defi:canonical_model_Ki}
A preferential model ${\enne}_i= \langle \Delta, <_i, \cdot^I \rangle$ of $K_i= \langle {\cal T}\cup m_i, {\cal A}\rangle$,
is a {\em lexicographic model of $K_i$} if  $\langle \Delta, \cdot^I \rangle$ is an $\alc$ model of $ \langle {\cal T}, {\cal A}\rangle$ 
and $<_i$ satisfies the following condition:
\begin{align}\label{cond:preference_prop_i}
x <_i y \mbox{\ \ {\bf \em iff} } V_i(x)  \prec_i V_i(y).
\end{align}
\end{definition}
Informally, $<_{C_j}$ gives higher preference to domain elements violating less typicality inclusions of $m_i$ with higher rank. 
In particular,  all  $x,y \not \in C_i^I$, $x \sim_{C_i} y$,
i.e., all $\neg C_i$-elements are assigned the same preference wrt $<_i$, the least one, as they trivially satisfy all the typicality properties in $m_i$.
As in Lehmann's semantics, 
in a lexicographic model ${\enne}_i= \langle \Delta, <_i, \cdot^I \rangle$ of $K_i$, the preference relation  $<_i$ is  
a strict {\em modular} partial order, i.e. an irreflexive, transitive and modular relation. 
As well-foundedness trivially holds for finite interpretations, a lexicographic model ${\enne}_i$ of $K_i$  is a ranked model of $K_i$.

\begin{proposition}
A lexicographic model ${\enne}_i = \langle \Delta, <_i, \cdot^I \rangle$ of $K_i= \langle {\cal T}\cup m_i, {\cal A}\rangle$  is a ranked model of $K_i$.
\end{proposition}

A multi-concept model for $K$ can be defined as a multi-preference interpretation with a preference relation $<_i$ for each module $m_i$.

\begin{definition}[Multi-concept interpretation]
Let $K= \mathit{\langle {\cal T}, {\cal D}, m_1, \ldots, m_k, {\cal A}, s \rangle}$ be a multi-concept knowledge base.
A {\em  multi-concept interpretation} $\emme$ for $K$ is a tuple $\langle \Delta, <_1, \ldots, <_k, \cdot^I \rangle$
such that, for all $i=1,\ldots, k$, $\langle \Delta, <_i, \cdot^I \rangle$ is a ranked $\alct$ interpretation, as defined in Section \ref{sec:ALC}.
\end{definition}

\begin{definition}[Multi-concept lexicographic model]
Let $K= \mathit{\langle {\cal T}, {\cal D}, m_1, \ldots, m_k, {\cal A}, s \rangle}$ be a multi-concept knowledge base.
A {\em multi-concept lexicographic model} $\emme=  \langle \Delta, <_1, \ldots, <_k, \cdot^I \rangle$ of $K$ is a  multi-concept interpretation for $K$,
such that,  
for all $i=1,\ldots,k$, 
${\enne}_i= \langle \Delta, <_i, \cdot^I \rangle$ is a  
lexicographic model of $K_i= \langle {\cal T}\cup m_i, {\cal A}\rangle$.
\end{definition}
 A canonical multi-concept lexicographic model of $K$  is multi-concept lexicographic model of $K$ such that $\Delta$ and $\cdot^I$ are the domain and interpretation function of some canonical preferential model of $ \langle {\cal T}\cup {\cal D}, {\cal A}\rangle$, according to Definition \ref{def-canonical-model-DL}.

\begin{definition}[Canonical multi-concept lexicographic model]
Given a multi-concept knowledge base  $K= \mathit{\langle {\cal T}, {\cal D}, m_1, \ldots, m_k, {\cal A}, s \rangle}$, 
a {\em canonical multi-concept lexicographic model} of $K$, $\emme=  \langle \Delta, <_1, \ldots, <_k, \cdot^I \rangle$, is 
 a  multi-concept lexicographic model of $K$ such that there
 is a canonical  $\alct$ model  $ \langle \Delta, <^*, \cdot^I \rangle$ of  $ \langle {\cal T}\cup {\cal D}, {\cal A}\rangle$, for some $<^*$.
 \end{definition}
%


Observe that,
restricting to the propositional fragment of the language (which does not allow universal and existential restrictions nor assertions), for a knowledge base $K$ without strict inclusions and
with a single module $m_1$, with subject $\top$, containing all the typicality inclusions in $K$,  the preference relation $<_1$ 
corresponds to Lehmann's lexicographic closure semantics, as its definition is based on the set of all defeasible inclusions in the knowledge base. 


\section{The combined lexicographic model of a KB} \label{sec:combination}

For multiple modules, each $<_i$ determines a ranked preference relation which can be used to answer {\em queries over  module $m_i$} (i.e. queries whose subject is $C_i$).
If we want to evaluate the query $\tip( C) \sqsubseteq D$  (are all typical $C$ elements also $D$ elements?) in module $m_i$ (assuming that  $C$ concerns  subject $C_i$), we can answer the query using the $<_i$ relation, by checking whether $min_{<_i}(C^I) \subseteq D^I$.
For instance, in Example  \ref{exa:student}, the query  ``are all typical Phd students young?" can be evaluated in module $m_2$. The answer would be positive, as the property of students of being normally young is inherited by PhD Student.  
The evaluation of a query in a specific module is something that is considered in context-based formalisms, such as in the CKR framework \cite{Bozzato14},
where there is a language construct $ \mathit{eval(X,c)}$ for evaluating a concept (or role) $X$ in context $c$.

The lexicographic orders $<_i$ and $<_j$ (for $i \neq j$) do not need to agree. For instance, in Example \ref{exa:student}, for two domain elements $x$ and $y$, we might have that $x <_1 y$ and $y <_2 x$, as $x$ is more typical than $y$ as an employee, but less typical than $x$ as a student.
To answer a query $\tip(C) \sqsubseteq D$, where $C$ is a concept which is concerned with more than one subject in the knowledge base (e.g., are typical employed students young?),
we need to {\em combine the relations $<_i$}.


A simple way of combining the modular partial order relations $<_i$ is to use Pareto combination.  
Let $\leq_i$ be defined as follows: $x \leq_i y$ iff $y \not <_i x$. As $<_i$ is a modular partial order, $\leq_i$ is a total preorder. 
Given a canonical multi-concept lexicographic model $\emme=  \langle \Delta, <_1, \ldots, <_k, \cdot^I \rangle$ of $K$, 
 we define a global preference relation $<$ on $\Delta$ as follows:
 \begin{align*} 
x <y  \mbox{\em \ \  iff \ \ }
(i) & \mbox{ for some }  i=1,\ldots,k, \;  x <_{i} y \mbox{ and } \ \ \ \ \ \ \ \ \ \ \ \  (*) \\
(ii) &  \mbox{ for all } j =1,\ldots,k ,  x \leq_j y,  
\end{align*}
The resulting relation $<$ is a partial order but, in general, modularity does not hold for $<$.
\begin{definition}
Given a canonical multi-concept lexicographic model $\emme=  \langle \Delta, <_1, \ldots, <_k, \cdot^I \rangle$ of $K$,
the {\em combined lexicographic interpretation of $\emme$}, is a triple $\emme^\Pe =\langle \Delta, <, \cdot^I \rangle$, where $<$ is the global preference relation defined by (*).
\end{definition}
We call $\emme^\Pe$ a {\em combined lexicographic model of $K$} (shortly, an $m^c_l$-model of $K$).
\begin{proposition}
A combined lexicographic model $\emme^\Pe$ of $K$ is a  preferential interpretation satisfying all the strict inclusions and assertions in $K$.
\end{proposition}

 A combined lexicographic model $\emme^\Pe$ of $K$ is a  preferential interpretation as those defined for $\alct$ in Definition \ref{semalctr}
 (and, in general, it is not a ranked interpretation).
However, preference relation $<$ 
 in $\emme^\Pe$ is not an arbitrary irreflexive, transitive and well-founded relation. It is obtained by first computing the lexicographic preference relations $<_i$ for modules, and then by combining them  into $<$.
As $\emme^\Pe$ satisfies all strict inclusions and assertions in $K$ but is not required to satisfy all typicality inclusions $\tip(C) \sqsubseteq  D$ in $K$,
 $\emme^\Pe$ is {\em not} a preferential $\alct$ model of $K$ as defined in Section \ref{sec:ALC}.

Consider a situation in which there are two concepts, $\mathit{Student}$ and $\mathit{YoungPerson}$, that are very related in that  students are normally young persons and young persons are normally students (i.e., $\mathit{\tip(Student) \sqsubseteq YoungPerson}$ and $\mathit{\tip(YoungPerson) \sqsubseteq Stu}$- 
 $\mathit{dent}$) and suppose there are two modules $m_1$ and $m_2$ such that $s(m_1)=\mathit{Student}$ and $s(m_2)=\mathit{YoungPerson}$.  The two classes may have different (and even contradictory) prototypical properties, for instance, normally students are quiet (e.g., when they are in their classrooms), $\mathit{\tip(Student) \sqsubseteq  Quiet}$, but normally young persons are not quiet,
 $\mathit{\tip(YoungPerson) \sqsubseteq  \neg Quiet}$. Considering the preference relations $<_1$ and $<_2$, associated with the two modules in a canonical multi-concept lexicographic model, we may have that, for  two young persons Bob and John, which are also students, $\mathit{bob <_1 john}$ and $\mathit{john <_2 bob}$, as Bob is quiet and John is not. Then, John and Bob are incomparable in the global relation $<$. Both of them, depending on the other  prototypical properties of students and young persons, might be minimal, among students, wrt the global preference relation $<$. Hence, the set  $\mathit{min_{<}(Student^I)} $ is not necessarily a subset of $\mathit{min_{<_1}(Student^I)} $.
 That is, typical students in the global relation may include instances (e.g., $\mathit{john}$) which do not satisfy all the typicality inclusions for $\mathit{Student}$, as they are are (globally) incomparable with the elements in $\mathit{min_{<_1}(Student^I)} $.
 This implies that the notion of $m^c_l$-entailment (defined below) cannot be stronger 
than preferential entailment in Section \ref{sec:ALC}. However, given the correspondence of $m^c_l$-models with the lexicographic closure in the case of a single module with subject $\top$, containing all the typicality inclusions in ${\cal D}$, $m^c_l$-entailment can neither be weaker than preferential entailment.
 
 In general, for a knowledge base $K$ and a module $m_i$, with $s(m_i)=C_i$, the inclusion $\mathit{min_{<}(C_i^I)} \subseteq \mathit{min_{<_i}(C_i^I)} $ may not hold and, for this reason, a combined lexicographic interpretation may fail to satisfy all typicality inclusions.
In this respect, canonical multi-concept lexicographic models are 
more liberal than KLM-style preferential models for typicality logics \cite{FI09}, where all the typicality inclusions are required to be satisfied and, in the previous example, 
$\mathit{min_{<}(Student^I) \subseteq Quiet^I}$ must hold for the typicality inclusion to be satisfied.
In fact, the knowledge base above is inconsistent in the preferential semantics and has no preferential model: from $\mathit{\tip(Student) \sqsubseteq Young}$- $\mathit{Person}$ and $\mathit{\tip(YoungPerson) \sqsubseteq Student}$, it follows that $\mathit{\tip(Student)= \tip(Young}$-  $\mathit{Person)}$ should hold in all preferential models of the knowledge base, which is impossible given the conflicting typicality inclusions $\mathit{\tip(Student) \sqsubseteq  Quiet}$ and  $\mathit{\tip(Young}$- 
$\mathit{Person) \sqsubseteq  \neg Quiet}$.

To require that all typicality inclusions in $K$ are satisfied in $\emme^\Pe$, the notion of $m^c_l$-model of $K$ can  be strengthened as follows.
\begin{definition}
A  {\em $\tip$-compliant $m^c_l$-model}  (or $m^c_l \tip$-model) $\emme^\Pe=\langle \Delta, <, \cdot^I \rangle$ of $K$ is a $m^c_l$-model  of $K$ such that all the typicality inclusions  in $K$ are satisfied in $\emme^\Pe$,
i.e., for all $\tip(C) \sqsubseteq  D \in {\cal D}$, 
$min_{<}(C^I) \subseteq D^I$.
\end{definition}
Observe that, $m^c_l \tip$-model $\emme^\Pe=\langle \Delta, <, \cdot^I \rangle$ of $K= \mathit{\langle {\cal T}, {\cal D}, m_1, \ldots, m_k, {\cal A}, s \rangle}$
is a KLM-style {\em preferential model} for the $\alct$ knowledge base $ \langle {\cal T}\cup {\cal D}, {\cal A}\rangle$,  
as defined in Section \ref{sec:ALC}.
As a difference, the preference relation $<$ in a  ${m^c_l} \tip$-model is not an arbitrary irreflexive, transitive and well-founded relation, but is defined from the lexicographic preference relations $<_i$'s according to equation (*). 

 We define a notion of {\em multi-concept lexicographic entailment ($m^c_l$-entailment)} in the obvious way: a query $F$ is $m^c_l$-entailed by $K$ ($K \models_{m^c_l} F$) if, for all $m^c_l$-models $\emme^\Pe =\langle \Delta, <, \cdot^I \rangle$ of $K$, $ F$ is satisfied in $\emme^\Pe$.
 Notice that a query $\tip(C) \sqsubseteq D$ is satisfied in  $\emme^\Pe$  when $min_<(C^I) \subseteq D^I$.
 
 Similarly, a notion of {\em $m^c_l \tip$-entailment} can be defined:  $K \models_{m^c_l \tip} F$ if, for all $m^c_l \tip$-models $\emme^\Pe =\langle \Delta, <, \cdot^I \rangle$ of $K$, $ F$ is satisfied in $\emme^\Pe$. 
 
 As, for any multi-concept knowledge base $K$,
  the set of $m^c_l \tip$-models of $K$ is a subset of the set of $m^c_l$-models of $K$, and there is some $K$ for which the inclusion is proper (see, for instance, the student and young person example above), 
   $m^c_l \tip$-entailment is stronger than $m^c_l$-entailment.
 It can be proved that both notions of entailment satisfy the KLM postulates of  preferential consequence relations, which can be reformulated for a typicality logic, considering that typicality inclusions $\tip(C) \sqsubseteq D$ \cite{lpar2007} stand for conditionals $C {\ent} D$ in KLM preferential logics \cite{KrausLehmannMagidor:90,whatdoes}.
See also \cite{BoothCasiniAIJ19} for the formulation of KLM postulates in the Propositional Typicality Logic (PTL).

In the following proposition, we let ``$\tip(C) \sqsubseteq D $" mean that $\tip(C) \sqsubseteq D $ is $m^c_l$-entailed from a given knowledge base $K$.


\begin{proposition} \label{prop:KLM_properties}
$m^c_l$-entailment 
satisfies the KLM postulates of  preferential consequence relations, namely:

{
\noindent
(REFL) \ $\tip(C) \sqsubseteq C $ \\
(LLE) \ If $A \equiv B$ and $\tip(A) \sqsubseteq C $, then $\tip(B) \sqsubseteq C $ \\
(RW) \  If $C \sqsubseteq D$ and $\tip(A) \sqsubseteq C $, then $\tip(A) \sqsubseteq D $ \\
(AND) \ If $\tip(A) \sqsubseteq C $ and $\tip(A) \sqsubseteq D $, then $\tip(A) \sqsubseteq C \sqcap D $\\
(OR) \ If $\tip(A) \sqsubseteq C $ and $\tip(B) \sqsubseteq C $, then $\tip(A \sqcup B) \sqsubseteq C $\\
(CM) \  If $\tip(A) \sqsubseteq D$ and $\tip(A) \sqsubseteq C $, then $\tip(A \sqcap D) \sqsubseteq C $ }
\end{proposition} 
Stated differently,  the set of the typicality inclusions $\tip(C) \sqsubseteq D $ that are $m^c_l$-entailed from a given  knowledge base $K$ 
is closed under conditions (REFL)-(CM) above. 
For instance, (LLE) means that if $A$ and $B$ are equivalent concepts in $\alc$ and $\tip(A) \sqsubseteq C $ is $m^c_l$-entailed from a given knowledge base $K$, than $\tip(B) \sqsubseteq C $  is also $m^c_l$-entailed from $K$; similarly for the other conditions (where inclusion $C \sqsubseteq D$ is entailed by $K$ in $\alc$).
It can be proved that also $m^c_l \tip$-entailment satisfies the KLM postulates of preferential consequence relations.

\normalcolor
 
 
%
It can be shown that both $m^c_l$-entailment  and $m^c_l \tip$-entailment  are not stronger than Lehmann's lexicographic closure in the propositional case.
Let us consider again Example \ref{exa:student}. 
 
\begin{example} \label{exa:HomeOwner}
Let us add another module $m_4$ with subject $Citizen$ to the knowledge base $K$, plus the following additional axioms in ${\cal T}$:

$\mathit{Italian \sqsubseteq Citizen}$  \ \ \ \ \ \ \ $\mathit{French \sqsubseteq Citizen}$ \ \  

$\mathit{Canadian \sqsubseteq Citizen}$


\noindent
Module $m_4$ has subject  $\mathit{Citizen}$, and contains the defeasible inclusions:

$(d_{17})$ $\mathit{\tip(Italian) \sqsubseteq  DriveFast}$ 


$(d_{18})$ $\mathit{\tip(Italian) \sqsubseteq  HomeOwner}$

\noindent
 Suppose  the following typicality inclusion is also added to module $m_2$: 

$(d_{19})$ $\mathit{\tip(PhDStudent) \sqsubseteq  \neg HomeOwner}$

\noindent
What can we conclude about typical Italian PhD students? 
We can see that  neither the inclusion $\mathit{\tip(PhDStudent \sqcap Italian ) \sqsubseteq  HomeOwner}$
nor the inclusion $\mathit{\tip(PhDStudent}$ $\mathit{ \sqcap Italian ) \sqsubseteq  \neg HomeOwner}$ are $m^c_l$-entailed by $K$.

In fact, in all canonical multi-concept lexicographic models $\emme=  \langle \Delta, <_1, \ldots, <_4, \cdot^I \rangle$ of $K$,
all elements in $min_{<_2}((PhDStudent \sqcap Italian)^I)$ ( the minimal Italian PhDStudent wrt $<_2$), 
have scholarship, are bright, are not home owners
(which are typical properties of PhD students),
have classes and are young  (which are properties of students not overridden for PhD students).

On the other end, all elements in $\mathit{min_{<_4}((PhDStudent }$ $\mathit{ \sqcap Italian)^I)}$ (i.e.,  the minimal Italian PhDStudent  wrt $<_4$) 
have the properties that they drive fast and are home owners. As $<_2$-minimal elements and $<_4$-minimal $\mathit{PhDStudent}$ $\mathit{ \sqcap Italian}$-elements are incomparable wrt $<$, the $<$-minimal Italian PhD students will include them all. \linebreak Hence, $\mathit{min_{<}((PhDStudent \sqcap Italian)^I) \not \subseteq HomeOwner^I}$ and $\mathit{min_{<}((PhDStudent \sqcap }$ \linebreak $\mathit{Italian)^I)  \not \subseteq (\neg HomeOwner)^I}$.


\end{example}
%
The home owner example is a reformulation of the example used by Geffner and Pearl  to show that the rational closure of conditional knowledge bases sometimes gives too strong conclusions, as  ``conflicts among defaults that should remain unresolved, are resolved anomalously" \cite{GeffnerAIJ1992}.
Informally, if defaults  $(d_{18})$ and $(d_{19})$ are conflicting  for Italian Phd students before adding any default  which makes PhD students exceptional wrt Students (in our formalization, default $(d_{10})$), they should remain conflicting after this addition. Instead, in the propositional case, 
both the rational closure \cite{whatdoes} and Lehmann's lexicographic closure \cite{Lehmann95} would entail that normally Italian Phd students are not home owners. This conclusion is unwanted, and is based on the fact that $(d_{18})$ has rank $0$, while $(d_{19})$ has rank $1$ in the rational closure ranking.
On the other hand, $\mathit{\tip(PhDStudent \sqcap Italian ) \sqsubseteq  \neg}$ $\mathit{ HomeOwner}$ is neither $m^c_l$-entailed from $K$,
nor  $m^c_l \tip$-entailed from $K$.
Both notions of entailment, when restricted to the propositional case, cannot be stronger than Lehmann's lexicographic closure.

Geffner and Pearl's Conditional Entailment \cite{GeffnerAIJ1992} does not suffer from the above mentioned problem as it is based on (non-ranked) preferential models.
The same problem, which is related to the representation of preferences as levels of reliability, has also been recognized by Brewka \cite{Brewka89} in his logical framework for default reasoning, 
leading to a generalization of the approach to allow a partial ordering between premises.
The example above shows that our approach using ranked preferences for the single modules, but a non-ranked global preference relation $<$ for their combination, 
does not suffer from this problem, provided a suitable modularization is chosen (in example above,  obtained by separating the typical properties of Italians and those of students in different modules).

\section{Further issues: Reasoning with a hierarchy of modules and user-defined preferences}

The approach considered in Section  \ref{sec:semantics} 
does not allow to reason with a hierarchy of modules, but it considers a flat collection of modules $m_1, \ldots, m_k$, each module concerning some subject $C_i$. As we have seen, a  module $m_i$ may contain defeasible inclusions referring to subclasses of $C_i$, such as $\mathit{PhDStudent}$  in the case of module $m_2$ with subject  $\mathit{Student}$.
When defining the preference relation $<_i$ the lexicographic closure semantics already takes into account the specificity relation among concepts within the module (e.g., the fact that $\mathit{PhDStudent}$ is more specific than $\mathit{Student}$).

However, nothing prevents us from defining two modules $m_i$ (with subject $C_i$) and $m_j$ (with subject $C_j$), such that concept $C_j$ is more specific than concept $C_i$. For instance, as a variant of Example \ref{exa:student}, we might have introduced two different modules $m_2$ with subject $\mathit{Student}$ and $m_5$ with subject $\mathit{PhDStudent}$. As concept $\mathit{PhDStudent}$  is more specific than concept $\mathit{Student}$ (in particular, $\mathit{PhDStudent \sqsubseteq Student}$ is entailed from the strict part of knowledge base ${\cal T}$ in $\alc$),
 the specificity information should be taken into account when combining the preference relations. 
 More precisely, preference $<_5$ should override preference $<_2$ when comparing $\mathit{PhDStudent}$-instances.


This is the principle followed by Giordano and Theseider Dupr\'{e} \cite{arXiv_ICLP2020} to define a global preference relation, in the case when each module with subject $C_i$  only contains  typicality inclusions of the form $\tip(C_i) \sqsubseteq D$.
A more sophisticated way to combine the preference relations $<_i$ into a global relation $<$ is used 
to deal with this case with respect to Pareto combination, by exploiting the specificity relation among concepts. 
While we refer therein for a detailed description of this more sophisticated notion of preference combination,  
let us observe that this solution could be as well applied to 
the modular multi-concept knowledge bases considered in this paper, provided an irreflexive and transitive notion of specificity among modules is defined. 

Another aspect that has been considered in the previously mentioned paper
is  the possibility of assigning ranks to the defeasible inclusions associated with a given concept.
 While assigning a rank to all typicality inclusions in the knowledge base may be awkward, often people have a clear idea about the relative importance of the properties for some specific concept.  For instance, we may know that the defeasible property that students are normally young is more important than the property that student normally do not have a scholarship. 
For small modules, which only contain typicality inclusions $\tip(C_i) \sqsubseteq D$ for a concept $C_i$, the specification of user-defined ranks of the $C_i$'s typical properties is a feasible option and a ranked modular preference relation can be defined from it, by using Brewka's  $\#$ strategy from his framework of Basic Preference Descriptions  for ranked knowledge bases \cite{Brewka04}. This alternative may coexist with the use 
of the lexicographic closure semantics built from the rational closure ranking for larger modules.
A mixed approach, integrating user-specified preferences with  the rational closure ranking for the same module, might be an interesting alternative.
This integration, however, does not necessarily provide a total preorder among typicality inclusions, which is our starting point for defining the modular preferences $<_i$ and their combination. Alternative semantic constructions should be considered for dealing with this case. 



According to the choice of fine grained or coarse grained modules, to the choice of the preferential semantics for each module (e.g., based on user-specified ranking or on Lehmann's lexicographic closure, or on the rational closure, etc.), and to the 
presence of a specificity relation among modules, alternative preferential semantics for modularized multi-concept knowledge bases can emerge.



   \section{Conclusions and related work}  \label{sec:conclu}

In this paper, we have proposed a modular multi-concept extension of the lexicographic closure semantics, based on the idea that defeasible properties in the knowledge base can be distributed in different modules, for which alternative preference relations can be computed. Combining multiple preferences into a single global preference allows a new preferential semantics and a notion of multi-concept lexicographic entailment ($m^c_l$-entailment) which, in the propositional case, is not stronger than the lexicographic closure. 

$m^c_l$-entailment satisfies the KLM postulates of a preferential consequence relation. 
It retains some good properties of the lexicographic closure, being able to deal with irrelevance, with specificity within the single modules,
and not being subject to the ``blockage of property inheritance" problem.
The combination of different preference relations provides a simple solution to a problem, recognized by Geffner and Pearl, that the rational closure of conditional knowledge bases sometimes gives too strong conclusions, as  ``conflicts among defaults that should remain unresolved, are resolved anomalously" \cite{GeffnerAIJ1992}. This  problem also affects the lexicographic closure, 
which is stronger than the rational closure.
Our approach using ranked preferences for the single modules, but a non-ranked preference $<$ for their combination, 
does not suffer from this problem, provided a suitable modularization is chosen.
As Geffner and Pearl's Conditional Entailment \cite{GeffnerAIJ1992}, 
also some non-monotonic DLs, such as
$\alctmin$, a typicality DL with a minimal model preferential semantics \cite{AIJ13},  
and the non-monotonic description logic ${\cal DL}^N$ \cite{bonattiAIJ15}, which supports normality concepts  based on a notion of overriding,   do not not suffer from the problem above.

Reasoning about exceptions in ontologies has led to the development of many non-monotonic extensions of Description Logics (DLs),  
incorporating non-monotonic features from most of NMR formalisms in the literature. 
In addition to those already mentioned in the introduction, let us recall the work by Straccia on inheritance reasoning in hybrid KL-One style logics \cite{Straccia93}
the work on defaults in DLs \cite{baader95a}, on description logics of minimal knowledge and negation as failure \cite{donini2002}, on circumscriptive DLs \cite{bonattilutz,Bonatti2011}, the generalization of rational closure to all description logics \cite{Bonatti2019}.
 as well as the combination of description logics and rule-based languages 
 \cite{Eiter2008,Eiter2011,rosatiacm,KnorrECAI12,Gottlob14,TPLP2016,Bozzato2018}.

Our multi-preference semantics is related with
 the multipreference semantics for $\alc$ developed by Gliozzi \cite{GliozziAIIA2016}, which is based on the idea of refining the rational closure construction considering the preference relations $<_{A_i}$ associated with different aspects, but we follow a different route concerning the definition of the preference relations associated with modules, and  the way of combining them in a single preference relation. In particular, 
 defining a refinement of rational closure semantics is not our aim in this paper,
as we prefer  to avoid some unwanted conclusions of rational and lexicographic closure while exploiting their good inference properties.

The idea of having different preference relations, associated with different typicality operators, has been studied by Gil \cite{fernandez-gil} to define a multipreference formulation of the typicality DL $\alctmin$, mentioned above. 
As a difference, in this proposal we associate preferences with modules and their subject,  
and we  combine the different preferences into a single global one. 
An extension of DLs with multiple preferences has also been developed by Britz and Varzinczak \cite{Britz2018,Britz2019} to define defeasible role quantifiers and defeasible role inclusions, by associating multiple preference relations with roles.

The relation of our semantics with the lexicographic closure for $\alc$ by Casini and Straccia \cite{casinistraccia2010,CasiniJAIR2013} should be investigated.  A major difference is in the choice of the rational closure ranking for $\alc$, but it would be interesting to check whether their construction corresponds to our semantics in the case of a single module $m_1$ with subject $\top$, when the same rational closure ranking is used.

Bozzato et al. present extensions of the CKR (Contextualized Knowledge Repositories) framework by Bozzato et al. \cite{Bozzato14,Bozzato2018} in which defeasible axioms are allowed in the global context 
and exceptions can be handled by overriding and have to be justified in terms of semantic consequence, considering sets of clashing assumptions for each defeasible axiom.
An extension of this approach to deal with general contextual hierarchies has been studied  by the same authors \cite{BozzatoES19}, by introducing a coverage relation among contexts, 
and defining a notion of preference among clashing assumptions, which is used to define a preference relation among justified {\em CAS} models, based on which CKR models 
are selected. 
An ASP based reasoning procedure, that is complete for instance checking, is developed for ${\cal SROIQ}$-RL.

For the lightweight description logic $\elpb$,  an Answer Set Programming (ASP) approach has been proposed  \cite{arXiv_ICLP2020} for defeasible inference in a miltipreference extension of $\elpb$, in the specific case in which each module only contains the defeasible inclusions $\tip(C_i) \sqsubseteq D$ for a single concept $C_i$, where the ranking of defeasible inclusions is specified in the knowledge base, following the approach by Gerhard Brewka in his framework of Basic Preference Descriptions  for ranked knowledge bases \cite{Brewka04}. A specificity relation among concepts is also considered. The ASP encoding exploits {\em asprin} \cite{BrewkaAAAI15}, by formulating multipreference entailment as a problem of computing preferred answer sets, 
which is  proved to be $\Pi^p_2$-complete. 
For $\elpb$ knowledge bases,
we aim at extending 
this ASP encoding to deal with the modular multi-concept  lexicographic closure semantics proposed in this paper, as well as with a more general framework, 
allowing for different choices of preferential semantics for the single modules and for different specificity relations for combining them. 
For lightweight description logics of the ${\cal EL}$ family \cite{rifel}, the ranking of concepts determined by the rational closure construction can be computed in polynomial time in the size of the knowledge base  \cite{SROEL_FI_2018,CasiniStracciaM19}.  
This suggests that we may expect a $\Pi^p_2$ upper-bound on the complexity of multi-concept lexicographic entailment.

\medskip

{\bf Acknowledgement:} We thank the anonymous referees for their helpful comments and suggestions. This research is partially supported by INDAM-GNCS Project 2019. 


\end{document}